\documentclass[letterpaper,twocolumn,10pt]{article}
\usepackage{usenix,epsfig,endnotes}
\usepackage[colorlinks=true,linkcolor=blue, citecolor=blue]{hyperref}
\usepackage{cite}
\usepackage{amsmath}
\usepackage{bbm}
\usepackage{dsfont}
\usepackage{bbold}
\usepackage{algpseudocode}
\usepackage{algorithm}

\newcommand{\scell}[2][c]{%
\begin{tabular}[#1]{@{}c@{}}#2\end{tabular}}
\usepackage{caption}

\usepackage[compact]{titlesec} 
\begin{document}
\date{}
\title{\Large \bf Abuse and Fraud Detection in Streaming Services \\ 
Using Heuristic-Aware Machine Learning}
\author{}
\author{
{\rm \normalsize Soheil Esmaeilzadeh}\\
\small Stanford Universtiy, CA\\
\small soes@alumni.stanford.edu
\and
{\rm \normalsize Negin Salajegheh}\\
\small Netflix, CA \\
\small nsalajegheh@netflix.com
\and
{\rm \normalsize Amir Ziai}\\
\small Netflix, CA \\
\small aziai@netflix.com
\and
{\rm \normalsize Jeff Boote}\\
\small Netflix, CA\\
\small jboote@netflix.com
}

\maketitle
\thispagestyle{empty}

\subsection*{Abstract}
This work presents a fraud and abuse detection framework for streaming services by modeling user streaming behavior. The goal is to discover anomalous and suspicious incidents and scale the investigation efforts by creating models that characterize the user behavior. We study the use of semi-supervised as well as supervised approaches for anomaly detection. In the semi-supervised approach, by leveraging only a set of authenticated anomaly-free data samples, we show the use of one-class classification algorithms as well as autoencoder deep neural networks for anomaly detection. In the supervised anomaly detection task, we present a so-called heuristic-aware data labeling strategy for creating labeled data samples. We carry out binary classification as well as multi-class multi-label classification tasks for not only detecting the anomalous samples but also identifying the underlying anomaly behavior(s) associated with each one. Finally, using a systematic feature importance study we provide insights into the underlying set of features that characterize different streaming fraud categories. To the best of our knowledge, this is the first paper to use machine learning methods for fraud and abuse detection in real-world scale streaming services. 

\section{Introduction}
\indent Today, streaming services serve content to millions of users all over the world. Most of these services allow users to stream or download content across a broad category of devices including mobile phones, laptops, and televisions; however, with some restrictions in place, such as the number of active devices, the number of streams, and the number of downloaded titles. So many users across so many platforms make for a uniquely large attack surface that includes content fraud, account fraud, and abuse of terms of service. Detection of fraud and abuse against streaming services at scale and in real-time is highly challenging. \\
\indent Data analysis and machine learning techniques are great candidates to help secure large-scale streaming platforms. Even though such techniques can scale security solutions proportional to the service size, they bring their own set of challenges such as requiring labeled data samples, defining effective features, and finding appropriate algorithms. In this work, by relying on the knowledge and experience of streaming security experts, we manually define more than 20 features based on the expected streaming behavior of the users and their interactions with devices. We present a systematic overview of the abusive streaming behaviors together with a set of model-based and data-driven anomaly detection strategies to identify them. \\
\indent The contributions of this work are as follows: (i) We define a set of anomalous behaviors that are common to subscription-based streaming platforms. (ii) We present a set of data featurization parameters that quantify the behavior of the clients of a streaming platform with the goal of identifying anomalous incidents. (iii) We present a wide range of semi-supervised and supervised (binary as well as multi-class multi-label) anomaly detection models and evaluate their performance for detecting anomalous incidents.\\
\indent The remainder of this paper is organized as follows: in subsections \eqref{sec:background_anomaly_detection} and \eqref{sec:streaming_platform}, we present a background on anomaly detection efforts and an overview of streaming platforms. In section \eqref{sec:datasetoverview}, we provide an overview of the dataset used in this work where (i) we describe the concept of heuristic as a data labeling strategy for the task of anomaly detection, (ii) we present an overview of the features, (iii) we present statistics of the dataset, and finally (iv) we explain our approach in tackling the label imbalance problem for the task of model-based anomaly detection. In section \eqref{sec:evaluation_metrics}, we provide an overview of the evaluation metrics that are used in this work to report the performance of the model-based anomaly detection approaches. In section \eqref{sec:model_based_anomaly_detection}, we describe a wide range of semi-supervised as well as supervised (binary as well as multi-class multi-label classifications) anomaly detection models. Finally, in section \eqref{sec:experiments}, we provide the results of our model-based anomaly detection efforts.

\subsection{Background on Anomaly Detection} \label{sec:background_anomaly_detection}
%
Anomalies (also known as outliers) are defined as certain patterns (or incidents) in a set of data samples that do not conform to an agreed-upon notion of normal behavior in a given context. Detecting anomalies is used in a broad range of applications such as fraud detection, identity theft, intrusion detection, surveillance applications, robots behaviors, and healthcare. Although anomaly detection has many applications, it faces a lot of challenges in practice. The precise definition of a normal (benign) behavior is often non-trivial and imprecise. This becomes particularly important when anomalies are the result of malicious activities by adversaries who adapt their signatures in order to appear benign and stay undetected. Moreover, in many applications, the notion of normality, as well as the important signatures required to characterize benign incidents, evolves over time requiring a constant evolution of anomaly detection strategies. Another challenging aspect is the fact that the definition of an anomalous incident is context-dependent, and this makes it impractical to use a universal anomaly detection model. Furthermore, in building data-driven anomaly detection models, the scarcity of reliably labeled data samples is a common challenge. Even if labeled data samples are available, since anomalous incidents occur far more rarely than normal events, the class imbalance becomes problematic. Another significant obstacle is data featurization and selecting correct and representative features.\\
\indent There are two main anomaly detection approaches, namely, (i) rule-based, and (ii) model-based. Rule-based anomaly detection approaches use a set of rules which rely on the knowledge and experience of domain experts. Domain experts specify the characteristics of anomalous incidents in a given context and develop a set of rule-based functions to discover the anomalous incidents \cite{1347773,helmer1998intelligent,salvador2004learning}. As a result of this reliance, the deployment and use of rule-based anomaly detection methods become prohibitively expensive and time-consuming at scale, and cannot be used for real-time analyses. Furthermore, the rule-based anomaly detection approaches require constant supervision by experts in order to keep the underlying set of rules up-to-date for identifying novel threats. Reliance on experts can also make rule-based approaches biased or limited in scope and efficacy. Accordingly, a fair amount of expertise, both in terms of depth and breadth, should be devoted to ensuring that a set of rules are put in place which could reliably detect a broad range of anomalous incidents over time. On the other hand, in model-based anomaly detection approaches, models are built and used to detect anomalous incidents in a fairly automated manner. Although model-based anomaly detection approaches are more scalable and suitable for real-time analysis, they highly rely on the availability of (often labeled) context-specific data. Model-based anomaly detection approaches, in general, are of three kinds, namely, (i) supervised, (ii) semi-supervised, and (iii) unsupervised. Supervised anomaly detection models require a set of labeled samples as anomalous and benign for training. Given a labeled dataset, a supervised anomaly detection model can be built to distinguish between anomalous and benign incidents. In semi-supervised anomaly detection models, only a set of benign examples are required for training. These models learn the distributions of benign samples and leverage that knowledge for identifying anomalous samples at the inference time. Finally,  unsupervised anomaly detection models do not require any labeled data samples, but it is not straightforward to reliably evaluate their efficacy.

\begin{figure*}[!h]
    \centerline{\includegraphics[width=0.45\paperwidth]{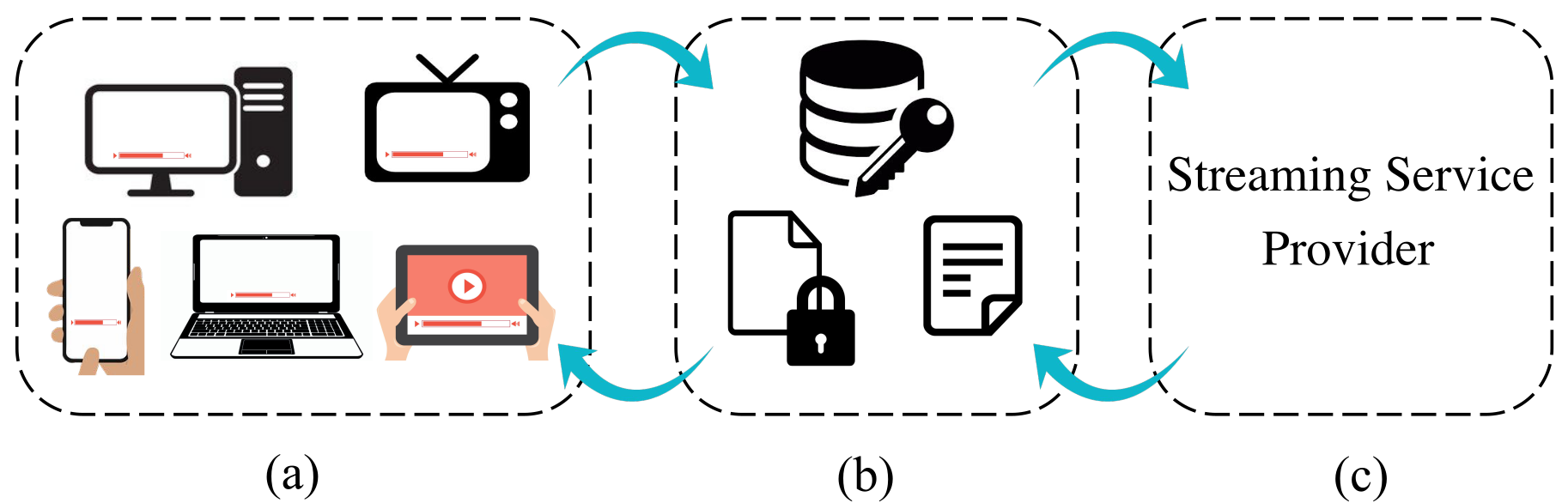}}
    \caption{Schematic of a streaming service platform: subfigure (a) illustrates device types that can be used by clients for streaming, subfigure (b) designates the set of authentication and authorization systems such as DRMs and license and manifest servers for providing encrypted contents as well as decryption keys and manifests, and subfigure (c) shows the streaming service provider, as a surrogate entity for digital content providers, that interacts with the other two components. }
    \label{fig:drm_schematic}
\end{figure*}

\subsection{Streaming Platforms}\label{sec:streaming_platform}

\noindent Commercial streaming platforms mainly rely on Digital Rights Management (DRM) systems. DRM is a collection of access control technologies that are used for protecting the copyrights of digital media such as movies and music tracks. DRM helps the owners of digital products prevent illegal access, modification, and distribution of their copyrighted work. DRM systems provide continuous content protection against unauthorized actions on digital content and restrict it to streaming and in-time consumption. The backbone of DRM is the use of digital licenses \cite{1463166}, which specify a set of usage rights for the digital content and contain the permissions from the owner to stream the content via an on-demand streaming service \cite{Callister2003}. These usage rights entail pieces of information such as frequency or duration of access, expiration date, restriction of transfer to other devices, copy or distribution permissions, resolution of the content, and can be used to enforce certain subscription-based business models. For instance, DRM could enforce that a specific movie title can only be viewed for a certain duration, on a specific device, with a particular resolution, and in a certain geographical location. Microsoft's PlayReady, Apple's  FairPlay, and Google's  Widevine are examples of commonly used DRM systems. The basics of digital rights management are common among different DRM types and involve four main components. With the content provider and the client being respectively the upstream and downstream components, there are two main parties involved in between, namely, the distributor, and the clearinghouse as the license server. At first, the content provider encodes the digital content into the format type that is supported by the DRM system. Afterward, the digital content becomes protected by getting encrypted and transferred to the appropriate content distributor, \textit{e.g.}, a streaming server, and the license that contains the content decryption keys is submitted to the clearinghouse. Finally, on the client's side, a request is sent to the streaming server to obtain the protected encrypted digital content. In order to stream the digital content, the user requests a license from the clearinghouse that verifies the user's credentials. Once a license gets assigned to a user, using a Content Decryption Module (CDM), the protected content gets decrypted and becomes ready for preview according to the usage rights enforced by the license \cite{Wang2003}. A decryption key gets generated using the license, which is specific to a certain movie title, can only be used by a particular account on a given device, has a limited lifetime, and enforces a limit on how many concurrent streams are allowed. Accordingly, a key cannot be captured and used to decrypt other movie titles for other clients and devices. In summary, DRM systems (i) provide encryption of the content in order to prevent unauthorized access, (ii) provide decryption key management, and (iii) deliver and manage the conditional access control of the clients \cite{Azad2010}. DRM systems, and in particular the license servers, also have access to the client information such as Internet Protocol (IP) addresses and geolocation, device identity, subscription plan, and types of devices. \\
\indent Another relevant component that is involved in a streaming experience is the concept of manifest. Manifest is a list of video, audio, subtitle, etc. which comes in the form of a few Uniform Resource Locators (URLs) that are used by the clients to get the movie streams. Manifest is requested by the client and gets delivered to the player before the license request, and it itemizes the available streams. Manifest also shares with the device the information about resolutions and bitrates that are available and then the device chooses a compatible stream. Moreover, the possibility of changing or customizing the manifest dynamically, at a per-client level, makes it possible to tailor the streaming experience.\\
\indent In this work, we devote the main attention to the communication between the clients of streaming service providers with the DRM systems, and the main focus is given to the license-time behaviors and the associated interactions between the clients and the license distributing servers.

\section{Dataset}\label{sec:datasetoverview}
\noindent In this part, we discuss different aspects of the dataset that is used in this work. First, we provide an overview of our proposed data labeling approach in the form of rule-based heuristics and a list of streaming fraud categories. Next, we discuss the data featurization strategy along with a list of features that are considered in this work. Afterward, we present the overall statistics of the dataset. Finally, we describe our strategy in tackling the label imbalance problem that is common to anomaly detection tasks.
%
\subsection{Data Labeling} \label{subsec:data_labeling}
%
\noindent With the constantly growing capacity of data acquisition systems, an immense amount of data becomes readily available for the rapidly expanding data-driven modeling tasks. Considering the fact that supervised as well as semi-supervised machine learning models highly rely on reliably labeled samples, data labeling is deemed a crucial yet challenging task, especially for real-world and industrial applications. \\
\indent The main approaches of labeling data samples are: (i) using a trained subject matter expert (SME) for hand-labeling the samples \cite{Dushkin2019}; (ii) using structural a priori domain-specific rule-based assumptions about the labels corresponding to the data samples \cite{Ratner2016}; (iii) employing weakly supervised learning \cite{Zhou2018} for instance when (a) a small subset of data is labeled, or (b) coarse-grained labels exist (inexact supervision), or (c) the given labels are not always ground truth (inaccurate supervision); (iv) applying the transfer learning (TL) technique to use the models that are already trained on a similar task in order to label the unlabeled data samples \cite{Sun2017}. Such techniques have been used for data labeling in different areas of research such as disease diagnosis by medical images \cite{Rousseau2011,1806.05233,esm}, speech recognition \cite{Hakkani-Tur2002}, object detection \cite{Rosenberg2002}, text classification \cite{Rosenberg2002,1904.00788}, remote sensing \cite{Huang2015} anomaly detection \cite{AbdullahAlMamun2018}, image classification, and video labeling \cite{Bondi2017}. Considering the data labeling approaches (i) to (iv) and acknowledging the fact that fully engaging the domain experts in the task of data labeling is labor-intensive, costly, and time-consuming, it is alternatively desired to automate the data labeling task and minimize the involvements by experts.\\
\indent For the task of anomaly detection in streaming platforms, as we have neither an already trained model nor any labeled data samples, we use the approach (ii), \textit{i.e.}, structural a priori domain-specific rule-based assumptions, for data labeling. Accordingly, we define a set of rule-based \textit{heuristics} used for identifying anomalous streaming behaviors of clients and label them as anomalous or benign. The fraud categories that we consider in this work are (i) content fraud, (ii) service fraud, and (iii) account fraud. With the help of security experts, we have designed and developed heuristic functions in order to discover a wide range of suspicious behaviors. 
We then use such heuristic functions for automatically labeling the data samples. The labels of the data samples could correspond to one or multiple incidents of the three aforementioned fraud categories. We then use the labeled samples to build machine learning models. Even though the heuristic functions can discover some anomalous incidents, they can never replace the machine learning models. With the use of machine learning models we can carry out anomaly detection at scale and in real-time, discover new anomalous patterns with sophistication levels beyond what heuristics can ever discover, do feature importance studies to gain insights into the streaming signature of users, capture the evolution of anomalous behaviors over time, and finally, identify novel unforeseen anomalous incidents in the future.\\
\indent Function (1) presents a pseudo-code for labeling anomalous accounts using the heuristic functions. Each heuristic function takes as its input the account-level streaming data and returns a binary label (\textit{i.e.}, anomalous or benign) indicating whether or not the user is considered anomalous with respect to that heuristic. We have built several dozen different in-house heuristics, each focusing on a specific anomalous streaming signature. Accordingly, each anomalous user gets tagged by one or more than one of the heuristics. Finally, we map the heuristic-level labels to the three aforementioned fraud categories, \textit{i.e.}, content fraud, service fraud, and account fraud. As mentioned before, such label mapping can result in data samples with labels corresponding to one or multiple fraud categories.\\
\begin{algorithm}[!h]
\footnotesize
\label{alg:anomaly_samples}
   \caption*{\textbf{Function 1}: \footnotesize{finding anomalous accounts using heuristic functions}}
   \hrule
    \begin{algorithmic}[1]
      \Function{labeling$\_$anomalies}{accounts$\_$data, accounts$\_$ids}
        \State tagged$\_$account$\_$ids$\_$and$\_$heuristics = $[\,\,]$
        \For{ ($\mathbf{X}, y$) \textbf{in} (accounts$\_$data, accounts$\_$ids)}
            \For{$h$ \textbf{in} \textbf{Heuristics}}
                \If {($h$($\mathbf{X}$).anomalous == True)}
                    \State  taggedAccountIDs$\_$heuristics += [($y$,\,$h$.name)]
                \Else
                    \State \textbf{continue}
                \EndIf
            \EndFor
        \EndFor
        \State \Return taggedAccountIDs$\_$heuristics
       \EndFunction
\end{algorithmic}
\end{algorithm}
%
\indent In order to label a set of benign (non-anomalous) accounts a group of vetted users that are highly trusted to be free of any forms of fraud is used. As an example of a heuristic function, in Function (2), we present a simple and intuitive heuristic that is based on the fact that households commonly watch only a couple of movies in a single day. This means that the majority of users are expected to acquire less than a dozen movie licenses per day. This heuristic by itself is not indicative of an anomalous behavior as highly active members and large households could end up acquiring hundreds of playback licenses within a couple of days. However, acquiring an egregiously high number of playback licenses in a single day can be an indication of anomalous activities. The pseudo-code snippet presented in Function (2) expresses such a heuristic that by relying on the number of acquired playback licenses tags accounts as anomalous or benign. In this example, we assume that the account-level statistics are calculated and the heuristic is simply a function that takes the account information and tags an account as anomalous or benign. The threshold is a hyperparameter that is set by the security experts and is commonly based on analyzing the historical data distributions.
%
\begin{algorithm}[!h]
\footnotesize
   \caption*{\textbf{Function 2}: \footnotesize{a heuristic for tagging accounts acquiring too many licenses in a day}}
   \hrule
    \begin{algorithmic}[1]
      \Function{many$\_$licenses$\_$heuristic}{accounts$\_$data, thresh}
    \State $\triangleright$ inputs: \\
    ~~~~~~~~~~~~~~~accounts data \\
    ~~~~~~~~~~~~~~~threshold of license counts per account in a day
    \State $\triangleright$ output: \\
    ~~~~~~~~~~~~~~~accounts acquiring more licenses than the threshold$\_$value
    \State tagged$\_$account$\_$ids = set(\,)
        \For{ ($\mathbf{X}, y$) \textbf{in} (accounts$\_$data, accounts$\_$ids)}
            \If {$\mathbf{X}$.license$\_$cnt $>$ thresh}
                \State tagged$\_$account$\_$ids.add($y$)
            \EndIf
        \EndFor
        \State \Return tagged$\_$account$\_$ids
       \EndFunction
\end{algorithmic}
\label{alg:benign_samples}
\end{algorithm}

Heuristics can be more complex and rely on streaming signatures that are beyond the account level statistics. For example, more complex heuristics can use some precomputed aggregated mappings that account for very complex streaming patterns. In summary, heuristics are a very powerful and intuitive way for security experts to characterize the occurrence of potential anomalies and could as well serve as a way of labeling data samples. Another upside to generating heuristics is that they can be highly informative for feature engineering. \\
\indent Next, for confidentiality reasons we only share three more examples of our in-house heuristics that we have used for tagging anomalous accounts:\\
(i) \textit{Rapid license acquisition}: a heuristic that is based on the fact that benign users usually watch one content at a time and it takes a while for them to move on to another content resulting in a relatively low rate of license acquisition. Based on this reasoning, we tag all the accounts that acquire licenses very quickly as anomalous.\\ 
(ii) \textit{Too many failed attempts at streaming}: a heuristic that relies on the fact that most devices stream without errors while a device, in trial and error mode, in order to find the ``right'' parameters leaves a long trail of errors behind. Abnormally high levels of errors are an indicator of a fraud attempt. \\
(iii) \textit{Unusual combinations of device types and DRMs}: a heuristic that is based on the fact that a device type (\textit{e.g.}, a browser) is normally matched with a certain DRM system (\textit{e.g.}, Widevine). Unusual combinations could be a sign of compromised devices that attempt to bypass security enforcements. \\
\indent For confidentiality reasons we do not share the threshold values used in the example heuristics provided above. It should be noted that the heuristics may not be completely accurate and they might wrongly tag accounts as anomalous (\textit{i.e.}, false-positive incidents), for example in the case of a buggy client or device. However, the heuristics work as a great proxy to embed the knowledge of security experts in tagging anomalous accounts. The presented results in section\,\eqref{sec:experiments} show that our in-house hand-picked set of heuristics are indeed effective.

\subsection{Data Featurization} \label{subsec:data_featurization}
%
\begin{table}[!h]
\centering
\def\arraystretch{1.4}
\resizebox{1\columnwidth}{!}{
\begin{tabular}{||c|l|l||}
\hline
{ \textbf{}} &
  { \textbf{feature name}} &
  \multicolumn{1}{l||}{{ \textbf{description}}}         \\ \hline\hline
f1  &  dev\_type\_a\_pct     & percentage use of type (a) devices by an account                   \\
f2  &  dev\_type\_b\_pct     & percentage use of type (b) devices by an account                   \\
f3  &  dev\_type\_c\_pct     & percentage use of type (c) devices by an account                   \\
f4  &  dist\_cdm\_cnt        & count of distinct CDM instances used associated with an account      \\
f5  &  dist\_cdm\_ver\_cnt   & count of distinct CDM versions used associated with an account       \\
f6  &  dist\_ip\_cnt         & count of distinct IPs associated with an account                     \\
f7  &  dist\_drm\_cnt        & count of distinct DRMs associated with an account                     \\
f8  &  dist\_enc\_frmt\_cnt  & count of distinct encoding formats associated with an account        \\
f9  &  dist\_dev\_id\_cnt    & count of distinct devices associated with an account                 \\
f10 &  dist\_dev\_cat\_cnt   & count of distinct device types associated with an account            \\
f11 & dist\_hour\_cnt        & count of distinct streaming hours associated with an account         \\
f12 & dist\_profile\_cnt     & count of distinct profiles associated with an account                \\
f13 & dist\_title\_cnt       & count of distinct titles associated with an account                  \\
f14 & drm\_type\_a\_pct      & percentage use of DRM type (a) by an account                       \\
f15 & drm\_type\_b\_pct      & percentage use of DRM type (b) by an account                       \\
f16 & drm\_type\_c\_pct      & percentage use of DRM type (c) by an account                       \\
f17 & drm\_type\_d\_pct      & percentage use of DRM type (d) by an account                       \\
f18 & end\_frmt\_a\_pct      & percentage use of encoding format (a) by an account                \\
f19 & end\_frmt\_b\_pct      & percentage use of encoding format (b) by an account                \\
f20 & end\_frmt\_c\_pct      & percentage use of encoding format (c) by an account                \\
f21 & end\_frmt\_d\_pct      & percentage use of encoding format (d) by an account                \\
f22 & expect\_msg\_pct       & percentage appearance of an account level message                     \\
f23 & license\_cnt           & count of content licenses associated with an account                 \\
\hline
\end{tabular}
}
\caption{The list of streaming related features with the suffixes $\_$pct and $\_$cnt respectively referring to \textit{percentage} and \textit{count}.}
\label{table:features}
\end{table}

\noindent In the dataset that we consider for this work, each sample consists of a cross-sectional set of 23 distinct continuous numerical non-categorical features gathered over a time window of a day which collectively characterize the daily streaming behavior of an account. The features considered in this work and their corresponding descriptions are presented in Table\,\eqref{table:features}. The features mainly belong to two distinct classes. One class accounts for the number of distinct occurrences of a certain parameter/activity/usage in a day. For instance, the dist$\_$title$\_$cnt feature characterizes the number of distinct movie titles streamed by an account. The second class of features on the other hand captures the percentage of a certain parameter/activity/usage in a day. For instance, the drm$\_$type\_a$\_$pct feature characterizes the percentage of type (a) DRMs among different DRM types that have been used by an account for streaming. Due to confidentiality reasons, we have partially obfuscated the features, for instance, dev$\_$type$\_$a$\_$pct, drm$\_$type$\_$a$\_$pct, and end$\_$frmt$\_$a$\_$pct are intentionally obfuscated and we do not explicitly mention devices, DRM types, and encoding formats.

\subsection{Data Statistics}\label{subsec:overall_statistics}
\begin{figure}[!h]
    \centerline{\includegraphics[width=\columnwidth]{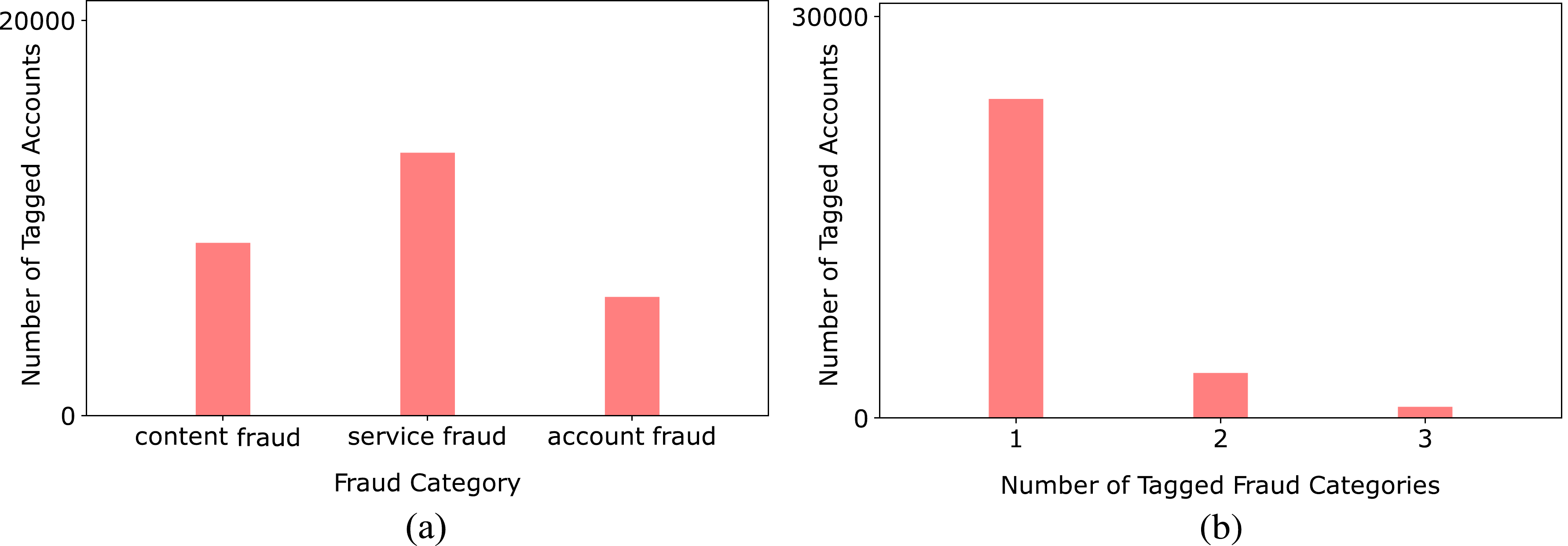}}
    \caption{Number of anomalous samples as a function of (a) fraud categories and (b) number of tagged categories.}
    \label{fig:dist_fraud_categories}
\end{figure}

\begin{figure*}[!h]
    \centerline{\includegraphics[width=0.5\paperwidth]{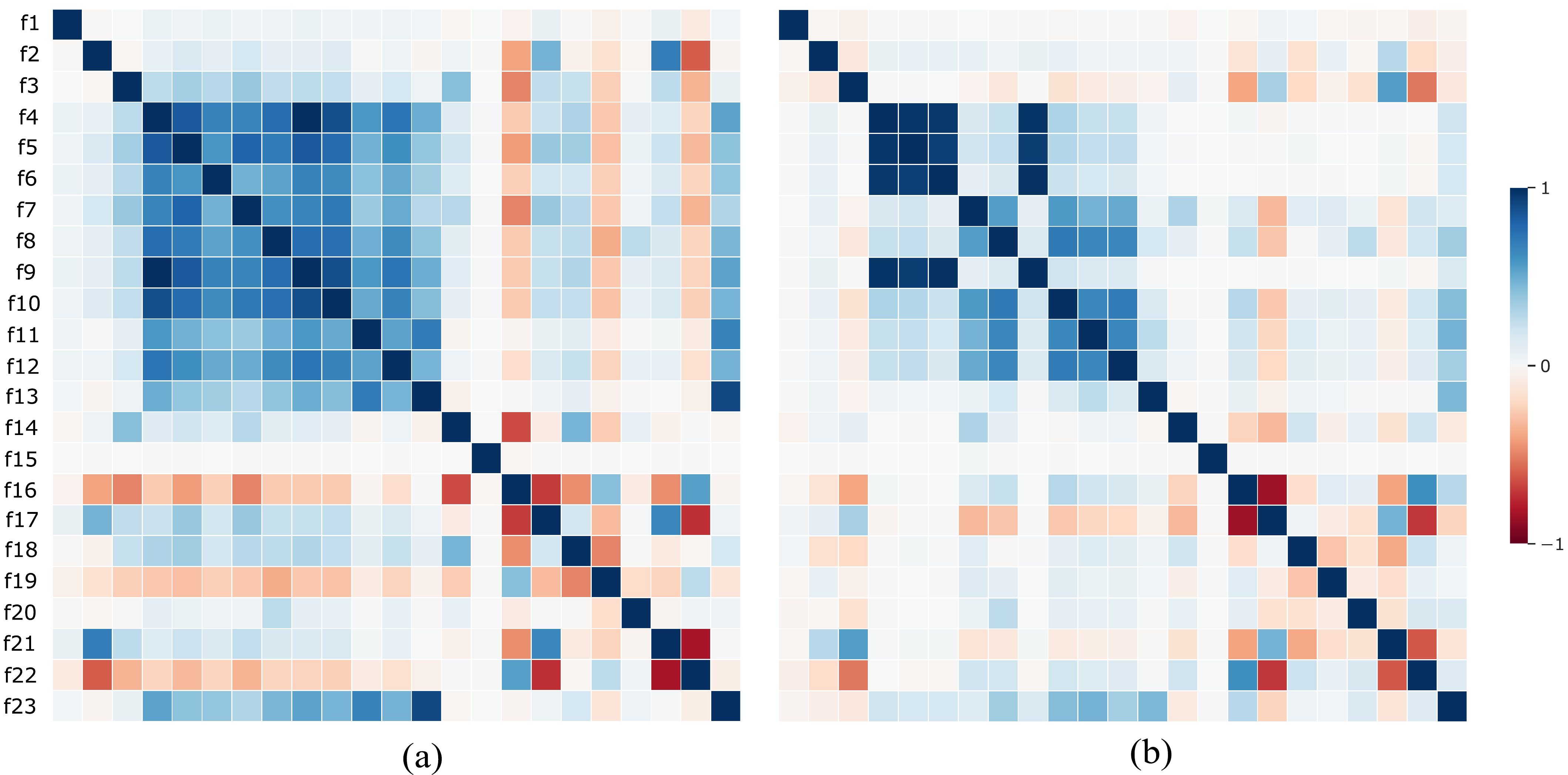}}
    \caption{Correlation matrix of the features presented in Table\,\eqref{table:features} for (a) clean and (b) anomalous data samples. }
    \label{fig:correlation_plots_all}
\end{figure*}

In this part, we present the statistics of the features presented in Table\,\eqref{table:features}. Over 30 days, we have gathered $1,030,005$ benign and $28,045$ anomalous accounts. The anomalous accounts have been identified (labeled) using the heuristic-aware approach presented in section \eqref{subsec:data_labeling}. Figure\,(\ref{fig:dist_fraud_categories}a) shows the number of anomalous samples as a function of fraud categories with $8,741\,(31\%)$, $13,299\,(47\%)$, $6,005\,(21\%)$ data samples being tagged as content fraud, service fraud, and account fraud, respectively. Figure\,(\ref{fig:dist_fraud_categories}b) shows that out of $28,045$ data samples being tagged as anomalous by the heuristic functions, $23,838\,(85\%)$, $3,365\,(12\%)$, and $842\,(3\%)$ are respectively considered as incidents of one, two, and three fraud categories. This means that only a small portion of the anomalous samples (\textit{i.e.}, only $3\%$) fall under all of the three fraud categories, whereas the majority of the anomalous samples (\textit{i.e.}, only $85\%$) belong to only one fraud category.\\
\indent Figure\,\eqref{fig:correlation_plots_all} presents the correlation matrix of the 23 data features described in Table\,\eqref{table:features} for clean and anomalous data samples. As we can see in Figure\,\eqref{fig:correlation_plots_all} there are positive correlations between features that correspond to device signatures, \textit{e.g.}, {dist$\_$cdm$\_$cnt} and {dist$\_$dev$\_$id$\_$cnt}, and between features that refer to title acquisition activities, \textit{e.g.}, {dist$\_$title$\_$cnt} and {license$\_$cnt}. We have carefully considered such correlations between the features and investigated the influence of removing or keeping the highly correlated features on the accuracy of machine learning based anomaly detection efforts presented in the following sections.
%
\subsection{Data Imbalance} \label{subsec:label_imbalance_treatment}
\begin{figure}[!h]
    \centerline{\includegraphics[width=0.37\paperwidth]{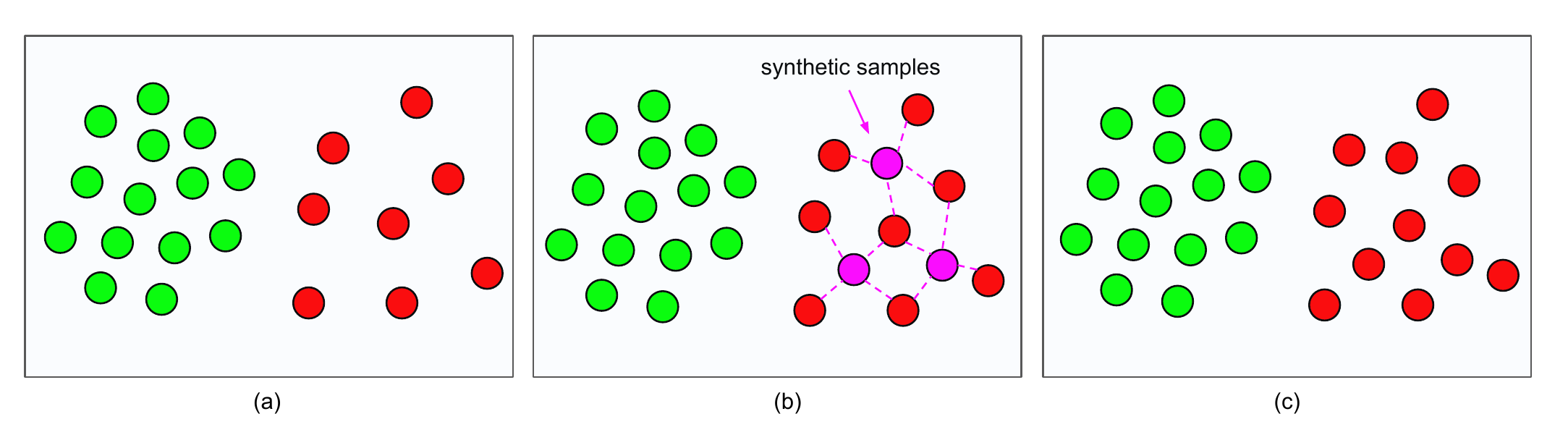}}
    \caption{Synthetic Minority Over-sampling Technique}
    \label{fig:SMOTE1}
\end{figure}

\indent Figure\,(\ref{fig:dist_fraud_categories}a) indicates a class imbalance as the anomalous tagged accounts are not equally distributed across the three fraud categories. It is well known that class imbalance can compromise the accuracy and robustness of the classification models. Accordingly, in this work, we use the Synthetic Minority Over-sampling Technique (SMOTE) \cite{Chawla2002,6235959} to over-sample the minority classes by creating a set of synthetic samples. 
Figure\,\eqref{fig:SMOTE1} shows a high-level schematic of Synthetic Minority Over-sampling Technique (SMOTE) with two classes shown in green and red where the red class has fewer number of samples present, \textit{i.e.}, is the minority class, and gets synthetically upsampled. 

\begin{figure}[!h]
    \centerline{\includegraphics[width=\columnwidth]{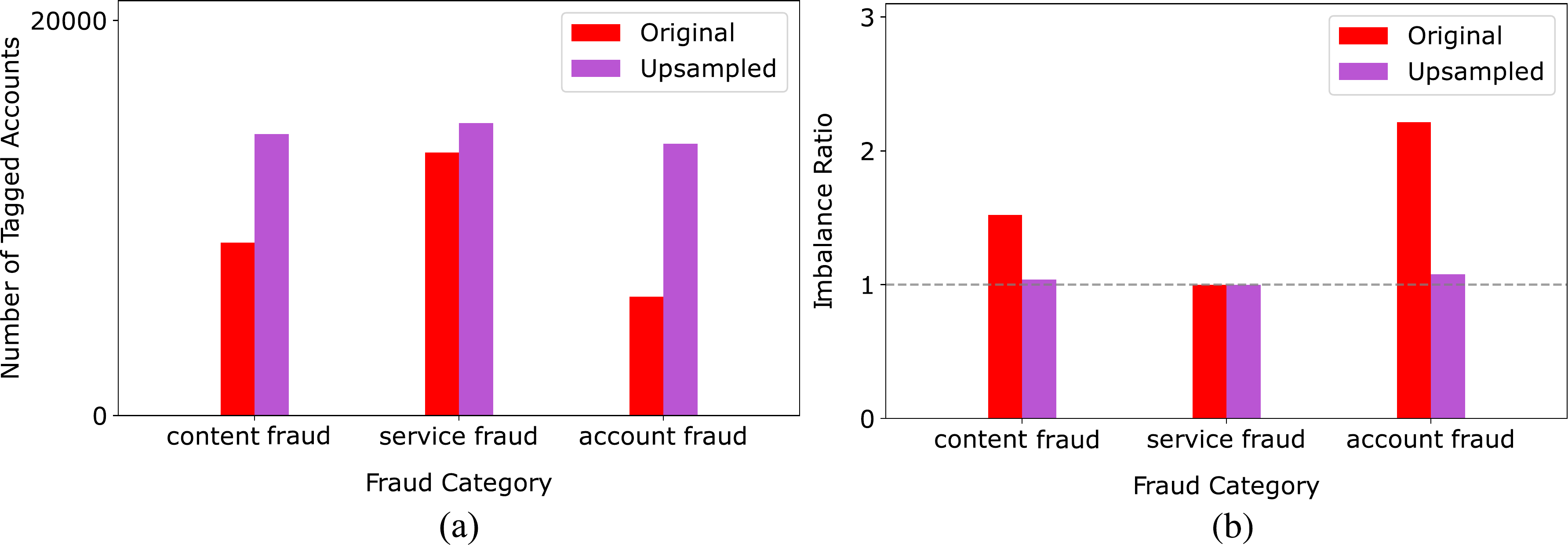}}
    \caption{For the three fraud categories before and after carrying out multi-class multi-label SMOTE: (a) number of anomalous tagged accounts and (b) label imbalance ratio. }
    \label{fig:upsampling_total}
\end{figure}

Function (3) in the Appendix section presents the pseudo-code for binary class SMOTE over-sampling. For the case of anomaly detection where there are multiple anomaly groups and one or more anomaly groups could be associated with a single account (\textit{i.e.}, multi-class multi-label anomaly detection) we use the Multi-label SMOTE (MLSMOTE) \cite{Charte2015,Giraldo-Forero2013} which is similar to the regular SMOTE, whereas the so-called label imbalance ratio (LIR) plays a role in its upsampling procedure. The label imbalance ratio (LIR) for a given label $i$ among $L$ different labels is defined as 
\small
\begin{equation}
    \text{LIR}_i = \frac{\max\,[\,{\text{count of label  }  j }\,]}{\text{count of label  }  i}\text{\,\,\,with\,\,\,} i,\,j\in\{1,\,\cdots,\,L\}, 
\end{equation}
\normalsize
where the minimum value of LIR is unity and the larger the value of LIR for a given label, the more imbalanced the dataset is with respect to that specific label. In a label-balanced dataset, the LIR would ideally be unity for all the labels. In this work, we carry out MLSMOTE on the multi-class multi-label dataset with a maximum critical threshold value of LIR$_{cr}\approx1$ for the imbalance ratio of the labels. Function (4) in the Appendix section shows the pseudo-code for MLSMOTE. Figures (\ref{fig:upsampling_total}a) and (\ref{fig:upsampling_total}b) respectively show the number of tagged anomalous accounts and the label imbalance ratio across the three fraud categories before and after carrying out MLSMOTE. As it can be seen in Figure\,(\ref{fig:upsampling_total}b) the two minority fraud categories of content fraud and account fraud have respectively label imbalance ratio values of around 1.5 and 2.2 before upsampling, whereas after carrying out upsampling the label imbalance ratio for all three fraud categories approaches unity.
%
%
\section{Evaluation Metrics} \label{sec:evaluation_metrics}

In this work, for evaluating the performance of the anomaly detection models we consider a set of evaluation metrics and report their values. For the binary anomaly detection task, such metrics are accuracy, precision, recall, f$_{0.5}$, f$_{1}$, and f$_{2}$ scores, and area under the curve of the receiver operating characteristic (ROC AUC). With TP, FP, and FN respectively being the true-positive, false-positive, and false-negative rates, the precision and recall are respectively expressed as 

\begin{subequations}
\begin{equation}
  \text{precision} = \frac{\text{TP}}{\text{TP} + \text{FP}},
\end{equation}  
\normalsize
and
\small
\begin{equation}
  \text{recall} = \frac{\text{TP}}{\text{TP} + \text{FN}}.
\end{equation}
\end{subequations}
Once the precision and recall values are found, the f$_{0.5}$, f$_{1}$, and f$_{2}$ scores can be calculated as
\small
\begin{equation}
    f_{\beta} = \frac{(1+\beta^2)\,\text{precision}\times\text{recall}}{\beta^2\,\text{precision}+\text{recall}}\,\,\,\text{with  } \beta \in \{0.5,1,2 \}.
\label{eq:fbeta}
\end{equation}
\normalsize
For the multi-class multi-label anomaly detection task where each sample can belong to one or more than one of the three fraud categories the definition of precision, recall, and f$_{\beta}$ are different. In a multi-class multi-label case assume that the associated labels with the sample $j$ be represented as a one-hot label vector $y_j=[l_1,l_2,l_3,...,l_c]$ with $c$ being the total number of considered classes and $l_i \in \{0,1\}$ being the one-hot encoding set for label $l_i$. As an example if sample $1$ only belongs to class $1$ and $c$ (a two-label sample), its corresponding one-hot label vector becomes $y_1 = [1,0,0,...,1]$. If $\hat{y}_j$ denotes the predicted label vector for a sample $j$ with the ground-truth label vector of $y_j$, the precision and recall can be expressed as 

\begin{subequations}
\begin{equation}
  \text{precision} = \frac{1}{N} \sum_{i=1}^{N} \frac{|y_i \cap \hat{y_i}|}{|y_i|},
\end{equation}  
and
\begin{equation}
  \text{recall} = \frac{1}{N} \sum_{i=1}^{N} \frac{|y_i \cap \hat{y_i}|}{|\hat{y}_i|},
\end{equation}
\end{subequations}
with $N$ being the total number of samples in the dataset. The numerator in the expressions for precision and recall is the number of labels in the predicted label vector ($\hat{y}$) that match those of the ground truth label vector ($y$). In precision, the ratio computes how many of the correctly predicted true labels are actually in the ground truth labels, whereas in recall the ratio gives the fraction of the ground truth labels that are predicted correctly. After evaluating the values of precision and recall, the f$_{\beta}$ score can be calculated using Equation\,\eqref{eq:fbeta}. In the multi-class multi-label task we consider five other evaluation metrics as well, namely, exact match ratio (EMR) score, Hamming loss, and Hamming score. The EMR simply quantifies the counts of the samples that all of their labels have been predicted correctly. Hamming loss is the ratio of wrong labels to the total number of labels, and Hamming score is a label-based accuracy expressed as 

\begin{equation}
  \text{Hamming Score} = \frac{1}{N} \sum_{i=1}^{N} \frac{|y_i \cap \hat{y_i}|}{|y_i \cup \hat{y_i}|}, 
\end{equation}
where $\hat{y}$ and $y$ are the predicted and ground truth label vectors, respectively.
%
\section{Model-Based Anomaly Detection} \label{sec:model_based_anomaly_detection}

\begin{figure}[!h]
    \centerline{\includegraphics[width=1\columnwidth]{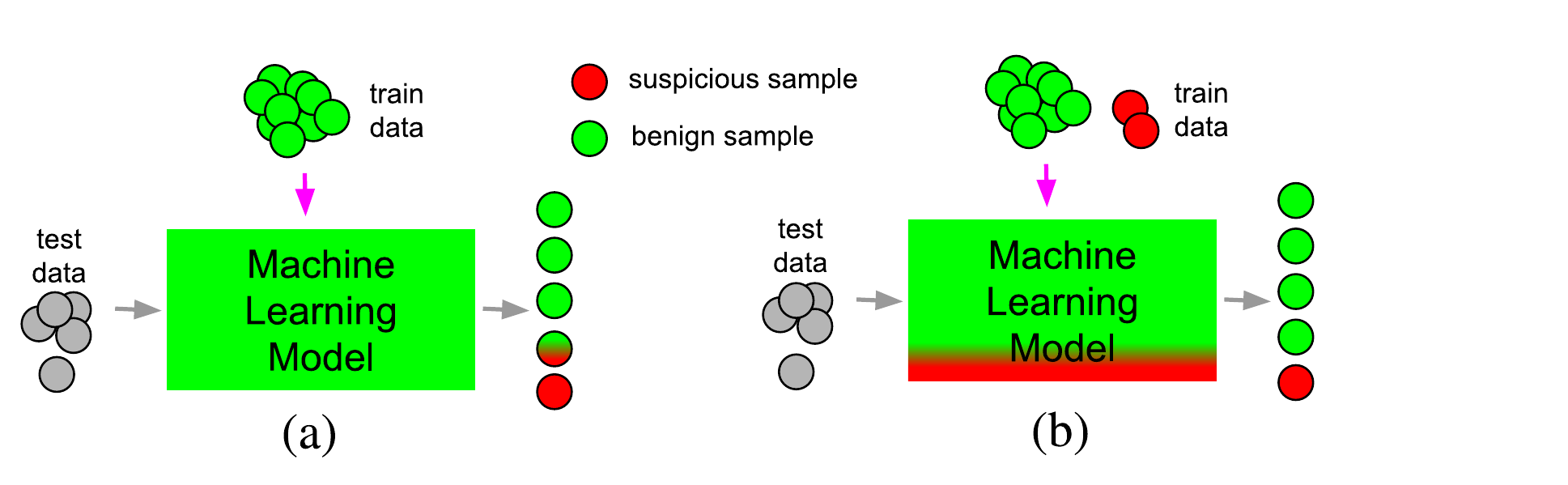}}
    \caption{Model-based anomaly detection approaches: (a) semi-supervised and (b) supervised.}
    \label{fig:modeling_overview}
\end{figure}

In this section, we briefly describe the modeling approaches that are used in this work for anomaly detection. We consider two model-based anomaly detection approaches, namely, (i) semi-supervised, and (ii) supervised as presented in Figure\,\eqref{fig:modeling_overview}. Given a set of $N$ data samples as $\mathbf{X} = \{\mathbf{x}_1,...,\mathbf{x}_N \}$ with $\mathbf{x}_i\,\in\,\mathbb{R}^D$ and $D$ being the number of features, a model-based anomaly detection can be considered as a mapping function $\Psi$ that projects the data samples onto a probability distribution or a target space comprised of classification labels, \textit{i.e.}, $\Psi$($\cdot$)\,:\,$\mathbf{X} \xrightarrow[]{}\,T$, and provides a distinction between the anomalous samples and the normal data instances.
%
\subsection{Semi-Supervised Anomaly Detection}
Semi-supervised anomaly detection approaches have the advantage of only requiring a benign set of data samples at the training stage, and are of interest in scenarios where reliably labeled anomalous samples are difficult to obtain, whereas reliably labeled benign samples exist. The key point about the semi-supervised model is that at the training step the model is supposed to learn the distribution of the benign data samples so that at the inference time it would be able to distinguish between the benign samples (that has been trained on) and the anomalous samples (that has not observed). Then at the inference stage, the anomalous samples would simply be those that fall out of the distribution of the benign samples.
\begin{figure}[!h]
    \centerline{\includegraphics[width=0.9\columnwidth]{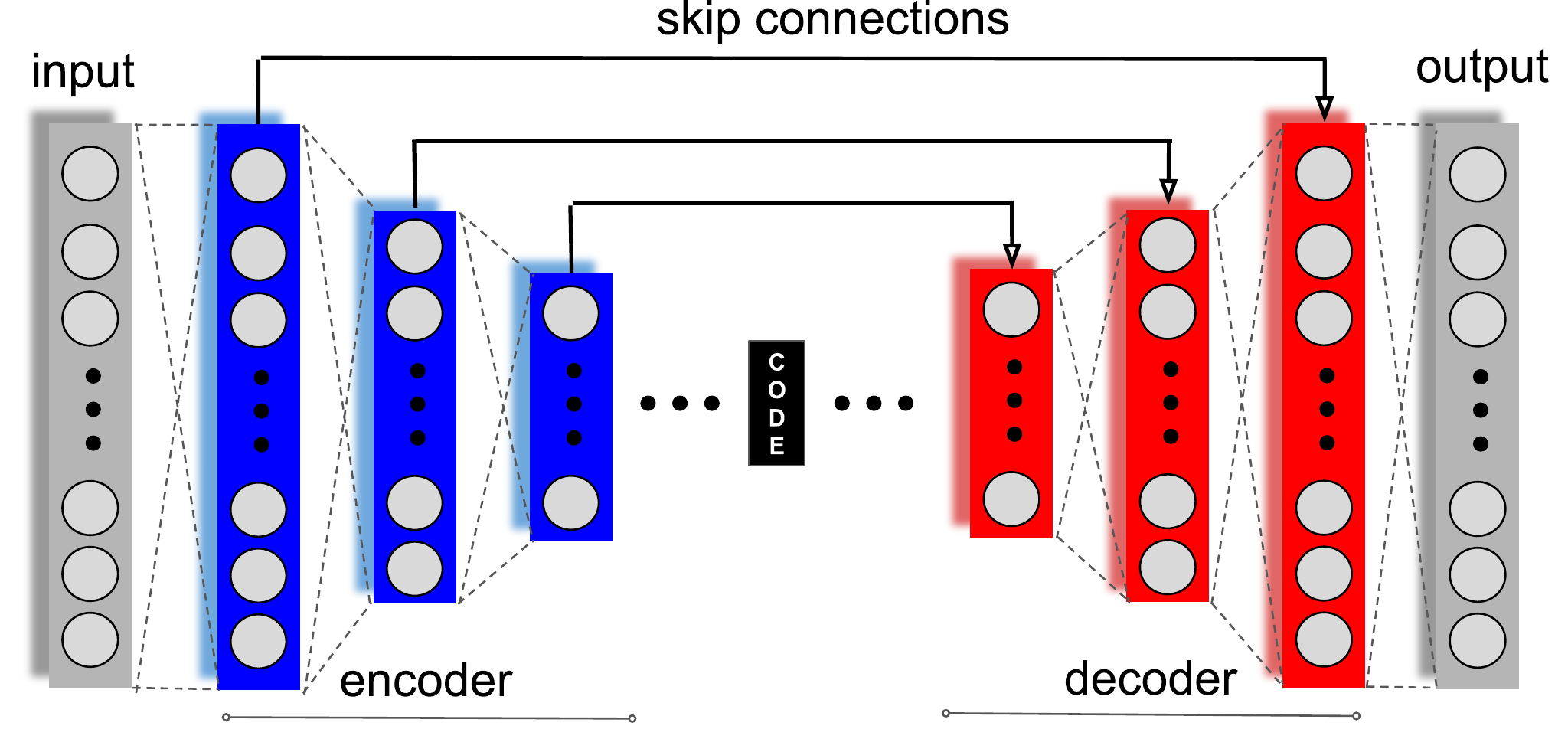}}
    \caption{Deep auto-encoder model as a semi-supervised anomaly detection approach. The encoding layers are followed by the decoding layers and skip connections connect the encoding and decoding stages. For training, the Mean Squared Error (MSE) of the input features and their reconstructed values (\textit{i.e.}, network's output) is used as the loss function.}
    \label{fig:autoencoder}
    \vspace{-5mm}
\end{figure}
\\
\indent The main approaches for semi-supervised anomaly detection are as follows. (i) One-Class methods such as One-Class Support Vector Machine (OC-SVM) \cite{1437839} which constructs a smooth boundary around the major probability mass of data; Isolation Forest (IF) \cite{Karczmarek2020} where a dataset is organized into a binary search tree and anomalies tend to be inserted at a lesser depth in a tree compared to benign samples; Elliptic Envelope (EE) \cite{Antonini2018} which generates an elliptical space around the central mass of the data in order to contain most of the normally distributed data; Local Outlier Factor (LOF) \cite{Ma2016,Cheng2019} which measures the local deviation of data; and (ii) deep neural autoencoders. The performance of One-Class methods could become sub-optimal when dealing with complex and high-dimensional datasets. However, supported by the literature, deep neural autoencoders can perform better than One-Class methods on complex and high-dimensional anomaly detection tasks. Deep neural autoencoders could be used either as a standalone feature extractor unit or in a hybrid manner where autoencoder extracts deep features and then feeds them to a separate anomaly detection unit such as an OC-SVM. However, since the hybrid approach extracts the deep features using an autoencoder and then feeds them into a separate anomaly detection unit, it fails to influence the representational learning in the hidden layers.\\
\indent In this work, as the One-Class anomaly detection approaches, we use the One-Class SVM, Isolation Forest, Elliptic Envelope, and Local Outlier Factor. As the deep neural autoencoder, we use a fully connected autoencoder shown in Figure\,\eqref{fig:autoencoder} where the encoding layers are followed by the decoding layers, and skip connections connect the two. We use the Mean Squared Error (MSE) between the input features and their reconstructed values (\textit{i.e.}, network's output) as the loss function during training and carry out hyperparameter tuning for the network architecture.
\subsection{Supervised Anomaly Detection}
Supervised anomaly detection approaches require fully labeled data comprised of both benign and anomalous samples for training. In this work, the dataset for the supervised anomaly detection is labeled using the heuristic functions as described in subsection \eqref{subsec:data_labeling}. Here, as the supervised anomaly detection, we carry out binary as well as multi-class multi-label classifications.
\subsubsection{Binary Classification.\, }
In the anomaly detection task using {binary} classification, we only consider two classes of samples namely benign and anomalous and we do not make distinctions between the types of the anomalous samples, \textit{i.e.}, the three fraud categories. For the binary classification task we use multiple supervised classification approaches, namely, (i) Support Vector Classification (SVC) approach \cite{Bi2005} which  finds the {best fit} hyperplane that divides and categorizes the data; (ii) K-Nearest Neighbors classification approach \cite{7898482} which is a non-parametric method that classifies an object by the majority vote of its neighbors; (iii) Decision Tree classification method \cite{Sharma2016} which builds a classification model in the form of a tree structure and breaks down a dataset into smaller subsets while at the same time develops an associated decision tree incrementally; (iv) Random Forest classification approach \cite{Cutler2012} which consists of multiple decision trees and uses a bagging strategy and feature randomness when building the individual trees; (v) Gradient Boosting \cite{Mason2000} and (vi) Ada-Boost Classifiers \cite{939503} which both are boosting algorithms that iteratively transform a set of weak learners into a strong learner; (vii) Nearest Centroid classification approach \cite{Tibshirani2003} which during classification assigns to a sample the label of the class of training samples with the mean (centroid) value closest to that sample; (viii) Quadratic Discriminant Analysis (QDA) classification approach \cite{4038449} which uses quadratic decision surfaces to separate different classes in a dataset; (ix) Gaussian Naive-Bayes classification strategy \cite{Abramson1963} which is a probabilistic classifier that is based on Bayes theorem and assumes a strong independence between the features; (x) Gaussian Process Classifier \cite{Ebden2008} which is a non-parametric classification method based on Bayesian inference that assumes some prior distribution about the probability densities of the data and its final classification is determined as the one that provides a good fit for the observed data; (xi) Label Propagation classification method \cite{Zhi2002} which relies on the proximity of labeled and unlabeled samples in the feature space for label assignments, and (xii) XGBoost \cite{xgboost} which is a scalable decision-tree-based ensemble algorithm that uses a gradient boosting framework. Finally, upon doing stratified k-fold cross-validation, we carry out an efficient grid search to tune the hyper-parameters in each of the aforementioned models for the binary classification task and only report the performance metrics for the optimally tuned hyper-parameters.
%
\subsubsection{Multi-Class Multi-Label Classification.\, }
%
In the anomaly detection task using multi-class multi-label classification, we consider the three fraud categories as the possible anomalous classes (hence multi-class), and each data sample is assigned one or more than one of the fraud categories as its set of labels (hence multi-label) using the heuristic-aware data labeling strategy presented in subsection \eqref{subsec:data_labeling}. For the multi-class multi-label classification task we use multiple supervised classification techniques, namely, (i) K-Nearest Neighbors, (ii) Decision Tree, (iii) Extra Trees, (iv) Random Forest, and (v) XGBoost. Similar to the binary classification part, upon doing stratified k-fold cross-validation, we carry out an efficient grid search to tune the hyper-parameters in each of the aforementioned models and report the performance metrics for the optimally tuned hyper-parameters. We also use the multi-class multi-label SMOTE upsampling technique presented in subsection \eqref{subsec:label_imbalance_treatment} to reduce the label imbalance ratio across the fraud categories and report the evaluation metrics for multi-class multi-label fraud detection on the original as well as the upsampled datasets.
%
\section{Results and Discussion} \label{sec:experiments}
\begin{figure*}[!h]
    \centerline{\includegraphics[width=0.8\paperwidth]{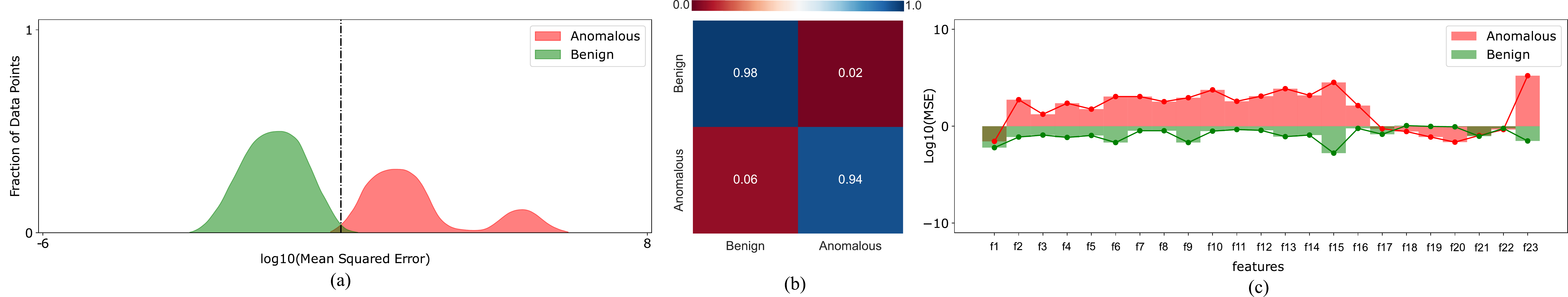}}
    \caption{For the deep auto-encoder model: (a) distribution of the Mean Squared Error (MSE) values for anomalous and benign samples at the inference stage - (b) confusion matrix across benign and anomalous samples - (c) Mean Squared Error (MSE) values averaged across the anomalous and benign samples for each of the 23 features.}
    \label{fig:distribution_confMatrix_autoEncoder_all}
\end{figure*}

\begin{figure*}[!h]
    \centerline{\includegraphics[width=0.75\paperwidth]{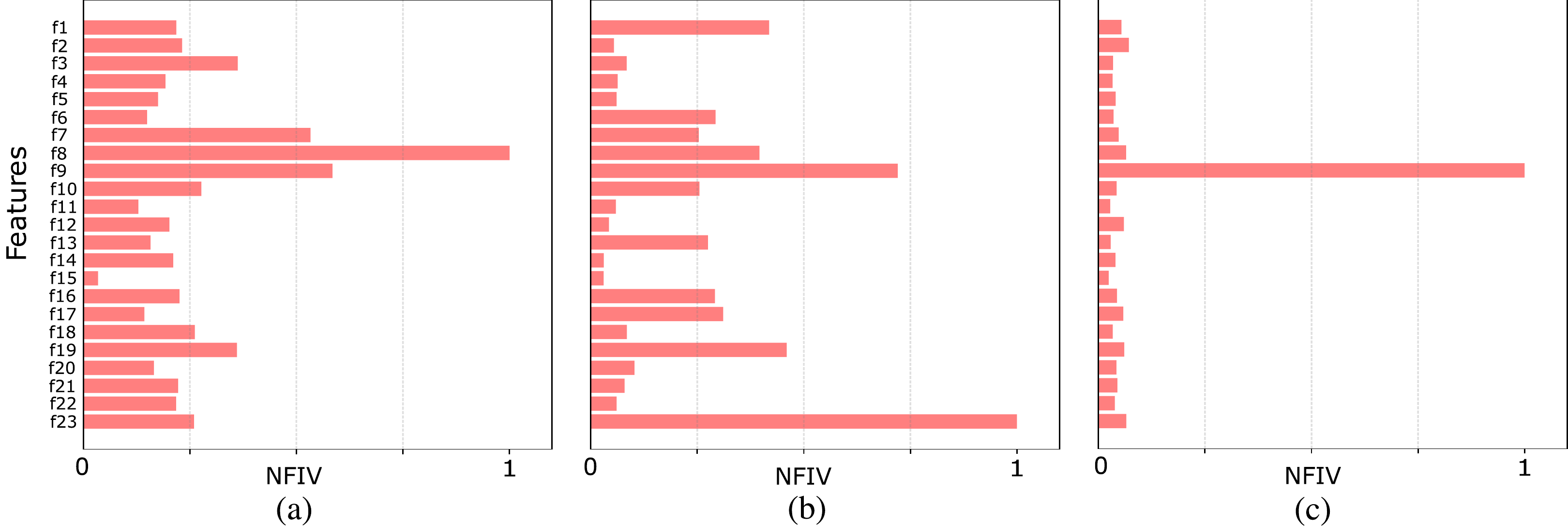}}
    \caption{The normalized feature importance values (NFIV) for the multi-class multi-label anomaly detection task using the XGBoost approach in Table\,\eqref{table:multiclass} across the three anomaly classes, \textit{i.e.}, (a) content fraud, (b) service fraud, and (c) account fraud.}
    \label{fig:feasture_importance_all}
\end{figure*}

In this part, we present the results of the model-based anomaly detection for semi-supervised and supervised approaches. Table\,\eqref{table:semi1} shows the values of the evaluation metrics for the semi-supervised anomaly detection methods, namely, One-Class Support Vector Machine, Isolation Forest, Elliptic Envelop, Local Outlier Factor, and deep auto-encoder.
\begin{table*}[!h]
\centering
\def\arraystretch{1}
\resizebox{0.6\paperwidth}{!}{
\begin{tabular}{||l|c|c|c|c|c|c|c||}
\hline
\multicolumn{1}{||c|}{ {   \textbf{Models}}}&
\multicolumn{1}{c|}{ {   \textbf{Accuracy}}}&
\multicolumn{1}{c|}{ {  \textbf{Precision}}}&
\multicolumn{1}{c|}{ {   \textbf{Recall}}} &
\multicolumn{1}{c|}{ {  \textbf{f}$_\mathbf{0.5}$ \textbf{score}}} &
\multicolumn{1}{c|}{ {  \textbf{f}$_\mathbf{1}$ \textbf{score}}} &\multicolumn{1}{c|}{ {   \textbf{f}$_\mathbf{2}$ \textbf{score}}} &
{   \textbf{ROC AUC}} \\ \hline\hline
One-Class SVM &0.237 & 0.997 & 0.161 & 0.488 & 0.277 & 0.193 & 0.58 \\
Isolation Forest & 0.093 & 1.000 & 0.002 & 0.010 & 0.004 & 0.002 & 0.50 \\
Elliptic Envelop & 0.307 & 0.997 & 0.239 & 0.610 & 0.385 & 0.281 & 0.62 \\
Local Outlier Factor &0.099 & 0.914 & 0.010 & 0.048 & 0.020 & 0.012 & 0.5 \\
Deep Auto-encoder & \textbf{0.966} & \textbf{0.964} & \textbf{0.935} & \textbf{0.941} & \textbf{0.949} & \textbf{0.953} & \textbf{0.96} \\
\hline
\end{tabular}
}
\caption{The values of the evaluation metrics for a set of semi-supervised anomaly detection models.}
\label{table:semi1}
\end{table*}
As we see from Table\,\eqref{table:semi1}, the deep auto-encoder model performs the best among the semi-supervised anomaly detection approaches with an accuracy of around $96\%$ and $\text{f}_1 \text{score}$ of $94\%$. Figure\,(\ref{fig:distribution_confMatrix_autoEncoder_all}a) shows the distribution of the Mean Squared Error (MSE) values for the anomalous and benign samples at the inference stage. The vertical dashed line is the threshold value for which the samples with smaller MSE values are considered benign and samples with larger MSE values are considered anomalous. We can see from Figure\,(\ref{fig:distribution_confMatrix_autoEncoder_all}a) that the trained deep auto-encoder model can well distinguish between the distribution of anomalous and benign samples. Figure\,(\ref{fig:distribution_confMatrix_autoEncoder_all}b) shows the confusion matrix for the deep auto-encoder model across the two classes of benign and anomalous samples. We can see that the deep auto-encoder model is confident in identifying the benign and anomalous samples with significantly low false-positive and false-negative rates. In Figure\,(\ref{fig:distribution_confMatrix_autoEncoder_all}c), for the deep auto-encoder model, we show the Mean Squared Error (MSE) values averaged across the anomalous and benign samples for each of the 23 features. We clearly see a distinction made by the deep auto-encoder model with respect to the features considering the two anomalous and benign classes.\\
%
\indent Table\,\eqref{table:binary} shows the values of the evaluation metrics for a set of supervised binary anomaly detection models, namely, Support Vector Classification, Nearest Centroid, k-Nearest Neighbors, Decision Tree, Random Forest, Label Propagation, Gaussian Process Classifier, Ada-Boos, Gaussian Naive-Bayes, Quadratic Discriminant Analysis, Gradient Boosting, and XGBoost. The Gradient Boosting approach outperforms the other classification models with accuracy and F-score values of around $86\%$. Moreover, we observe that in the binary anomaly detection task the Gradient Boosting approach as the best performing classifier achieves a balanced performance with respect to precision and recall.\\
\begin{table*}[!h]
\centering
\def\arraystretch{1}
\resizebox{0.7\paperwidth}{!}{
\begin{tabular}{||l|c|c|c|c|c|c|c||}
\hline
\multicolumn{1}{||c|}{ {   \textbf{Models}}}&
\multicolumn{1}{c|}{ {   \textbf{Accuracy}}}&
\multicolumn{1}{c|}{ {  \textbf{Precision}}}&
\multicolumn{1}{c|}{ {   \textbf{Recall}}} &
\multicolumn{1}{c|}{ {  \textbf{f}$_\mathbf{0.5}$ \textbf{score}}} &
\multicolumn{1}{c|}{ {  \textbf{f}$_\mathbf{1}$ \textbf{score}}} &\multicolumn{1}{c||}{ {   \textbf{f}$_\mathbf{2}$ \textbf{score}}} &
{   \textbf{ROC AUC}} \\ \hline\hline
Support Vector Classification & 0.45& 0.63 & 0.11 & 0.32 & 0.19 & 0.13 & 0.43 \\
Nearest Centroid & 0.56&0.85&0.14&0.43&0.24&0.17&0.56 \\
k-Nearest Neighbors & 0.78&0.82&0.71&0.8&0.76&0.73&0.78 \\
Decision Tree &0.82&0.81&0.82&0.81&0.82&0.82&0.81 \\
Random Forest & 0.8&0.8&0.81&0.8&0.81&0.81&0.8 \\
Label Propagation &0.77&0.81&0.7&0.79&0.75&0.72&0.78 \\
Gaussian Process Classifier &0.77&0.8&0.72&0.78&0.76&0.73&0.76 \\
Ada-Boost & 0.83&0.83&0.82&0.83&0.82&0.82&0.84 \\
Gaussian Naive-Bayes & 0.72&0.87&0.51&0.77&0.65&0.56&0.72 \\
Quadratic Discriminant Analysis & 0.72&\textbf{0.86}&0.52&0.76&0.65&0.56&0.72 \\
Gradient Boosting & \textbf{0.86}&0.85&\textbf{0.88}&\textbf{0.86}&\textbf{0.87}&\textbf{0.88}&\textbf{0.86} \\
XGBoost &0.84&0.83&0.84&0.83&0.84&0.85&0.84 \\
\hline
\end{tabular}
}
\caption{The values of the evaluation metrics for a set of supervised binary anomaly detection classifiers.}
\label{table:binary}
\end{table*}
\indent Table\,\eqref{table:multiclass} shows the values of the evaluation metrics for a set of supervised multi-class multi-label anomaly detection models, namely, k-Nearest Neighbors, Decision Trees, Extra Trees, Random Forest, and XGBoost. The values of evaluation metrics provided in parentheses correspond to the classifiers that are trained on the original (not upsampled) dataset. The XGBoost approach outperforms the other classification models with accuracy and F-score values of around $72\%$ on the upsampled dataset and around $65\%$ on the original (not upsampled) dataset. Moreover, we see that in the multi-class multi-label anomaly detection task the XGBoost approach as the best performing classifier achieves a balanced performance with respect to precision and recall.\\
\begin{table*}[!h]
\centering
\def\arraystretch{1.3}
\resizebox{0.7\paperwidth}{!}{
\begin{tabular}{||l|c|c|c|c|c|c|c|c|c||}
\hline
\multicolumn{1}{||c|}{  {   \textbf{Models}}} &
\multicolumn{1}{c|}{  {   \textbf{Precision}}} &
\multicolumn{1}{c|}{  {   \textbf{Recall}}} &
\multicolumn{1}{c|}{  {   \textbf{f}$_\mathbf{0.5}$ \textbf{score}}} &
\multicolumn{1}{c|}{  {   \textbf{f}$_\mathbf{1}$ \textbf{score}}} &
\multicolumn{1}{c|}{  {   \textbf{f}$_\mathbf{2}$ \textbf{score}}} &
\multicolumn{1}{c|}{  {   \textbf{ROC AUC}}} &
\multicolumn{1}{c|}{  {   \textbf{EMR}}} &
\multicolumn{1}{c|}{  {   \begin{tabular}[c]{@{}c@{}}\textbf{Hamming} \textbf{Loss}\end{tabular}}} &
\multicolumn{1}{c||}{  {   \begin{tabular}[c]{@{}c@{}}\textbf{Hamming} \textbf{Score}\end{tabular}}}  \\ \hline\hline
k-Nearest Neighbors & \scell{0.58\,(0.49)}& \scell{0.60\,(0.51)}& \scell{0.59\,(0.50)}& \scell{0.59\,(0.49)}& \scell{0.58\,(0.49)}& \scell{0.77\,(0.79)}& \scell{0.56\,(0.45)}& \scell{0.12\,(0.15)}& \scell{0.58\,(0.48)} 
\\ 
Decision Tree  & 
\scell{0.69\,(0.60)}& \scell{0.64\,(0.57)}& \scell{0.64\,(0.56)}& \scell{0.65\,(0.57)}& \scell{0.67\,(0.58)}& \scell{0.81\,(0.76)}& \scell{0.55\,(0.45)}& \scell{0.12\,(0.15)}& \scell{0.63\,(0.54)}
\\ 
Extra Trees & \scell{0.65\,(0.57)}& \scell{0.63\,(0.58)}& \scell{0.64\,(0.58)}& \scell{0.64\,(0.57)}& \scell{0.63\,(0.56)}& \scell{0.81\,(0.77)}& \scell{0.62\,(0.55)}& \scell{0.09\,(0.11)}& \scell{0.63\,(0.57)}
\\
Random Forest & \scell{0.57\,(0.52)}& \scell{0.59\,(0.54)}& \scell{0.58\,(0.53)}& \scell{0.57\,(0.52)}& \scell{0.57\,(0.52)}& \scell{0.77\,(0.74)}& \scell{0.54\,(0.48)}& \scell{0.102\,(0.12)}& \scell{0.56\,(0.51)}
\\ 
XGBoost & \scell{\bf{0.72}\,\bf{(0.65)}}& \scell{\bf{0.73}\,\bf{(0.65)}}& \scell{\bf{0.73}\,\bf{(0.64)}}& \scell{\bf{0.72}\,\bf{(0.64)}}& \scell{\bf{0.73}\,\bf{(0.65)}}& \scell{\bf{0.85}\,\bf{(0.80)}}& \scell{\bf{0.69}\,\bf{(0.61)}}& \scell{\bf{0.08}\,\bf{(0.11)}}& \scell{\bf{0.71}\,\bf{(0.63)}}
\\ 
\hline
\end{tabular}
}
\caption{The values of the evaluation metrics for a set of supervised multi-class multi-label anomaly detection approaches. The values in parenthesis refer to the performance of the models trained on the original (not upsampled) datasets.}
\label{table:multiclass}
\end{table*}
\indent Figures (\ref{fig:feasture_importance_all}a) to (\ref{fig:feasture_importance_all}c) show the normalized feature importance values (NFIV) for the multi-class multi-label anomaly detection task using the XGBoost approach, as the best performing classifier in Table\,\eqref{table:multiclass}, across the three anomaly classes, \textit{i.e.}, content fraud, service fraud, and account fraud. In Figure\,(\ref{fig:feasture_importance_all}a), for the content fraud category, the three most important features are the count of distinct encoding formats ({dist\_enc\_frmt\_cnt}), the count of distinct devices ({dist\_dev\_id\_cnt}), and the count of distinct DRMs ({dist\_drm\_cnt}). This implies that for content fraud the uses of multiple devices, as well as encoding formats, stand out from the other features. For the service fraud category in Figure\,(\ref{fig:feasture_importance_all}b) we see that the three most important features are the count of content licenses associated with an account ({license\_cnt}), the count of distinct devices ({dist\_dev\_id\_cnt}), and the percentage use of type (a) devices by an account (dev$\_$type$\_$a$\_$pct). This shows that in the service fraud category the counts of content licenses and distinct devices of type (a) stand out from the other features. Finally, for the account fraud category in Figure\,(\ref{fig:feasture_importance_all}c), we see that the count of distinct devices ({dist\_dev\_id\_cnt}) dominantly stands out from the other features.

\section{Conclusion and Future Work} \label{sec:conclusion_and_future_work}

In this work, we presented a framework for discovering anomalous user behaviors in streaming platforms. We introduced the concept of heuristics that leverages the knowledge of experts to label data for building machine learning models when labeled samples are missing. We introduced three streaming fraud categories that entail abnormal incidents regarding digital content, streaming service, and user account. We presented a multi-class multi-label synthetic minority over-sampling technique to circumvent the data imbalance problem in multi-class multi-label anomaly detection tasks. We studied the use of a wide range of semi-supervised and supervised machine learning models for the task of anomaly detection. We showed that the use of a deep auto-encoder model trained only on benign samples can achieve relatively high accuracy for binary class anomaly detection. We also found out that Gradient Boosting and XGBoost achieve the highest accuracy and F-score values for detecting anomalous incidents respectively for binary and multi-class multi-label scenarios. We further showed that the accuracy of the XGBoost classifier can further increase upon upsampling the multi-class multi-label set of anomalous samples. Finally, we presented a feature importance study for multi-class multi-label anomaly detection across the three fraud categories and provided insights into the correlation between the features and the streaming fraud categories. As the future directions, we plan to explore the potential of using the one-class models as feature extractors together with multi-class classifiers. We plan to also investigate the temporal representation of streaming activities of the users and explore the possibility of embedding time dependency into the data features.

\section*{Appendix} \label{sec:appendix}

Function (3) presents the pseudo-code for synthetic minority over-sampling technique (SMOTE) for binary class data samples.
\begin{algorithm}[!h]
  \footnotesize
   \caption*{\textbf{Function 3}: \footnotesize{SMOTE upsampling for binary classification}}
   \hrule
    \begin{algorithmic}[1]
    \State $\triangleright$ N: SMOTE percentage
    \State $\triangleright$ k: Number of nearest neighbors
    \State $\triangleright$ D := $\mathbf{X}_i$ with $i$ = $1,2,...,T$: Minority data
    \State $\triangleright$ T: Number of minority instances
      \Function{Binary$\_$SMOTE}{N, k, D}
      \State upsampled$\_$samples = $[\,\,]$
        \For{ $i$ = $1, 2, ..., $T}
            \State Find the $k$ nearest (minority class) neighbors of $\mathbf{X}_i$
            \State $\hat{\text{N}}$ = N$//100$
            \While {$\hat{\text{N}} \neq 0$}
            \State $\bar{\mathbf{X}}$ $\xleftarrow{}$ select one of the $k$ nearest neighbors
            \State $\alpha \in [0,1]$ $\xleftarrow{}$ select a random number
            \State $\hat{\mathbf{X}}$ = $\mathbf{X}_i + \alpha(\bar{\mathbf{X}}-\mathbf{X}_i)$ 
            \State upsampled$\_$samples += $\hat{\mathbf{X}}$
            \State $\hat{\text{N}}$ = $\hat{\text{N}} - 1$
            \EndWhile
        \EndFor
        \State \Return upsampled$\_$samples
       \EndFunction
\end{algorithmic}
\label{alg:SMOTE_binary1}
\end{algorithm}

Function (4) presents the pseudo-code for synthetic minority over-sampling technique (SMOTE) for multi-class multi-label data samples.
\begin{algorithm}[!h]
\footnotesize
   \caption*{\textbf{Function 4}: \footnotesize{SMOTE upsampling for multi-class multi-label classification}}
   \hrule
    \begin{algorithmic}[1]
    \State $\triangleright$ {k: Number of nearest neighbors}
    \State $\triangleright$ D: Multi-class multi-label dataset
    \State $\triangleright$ LIR$_{cr}$: maximum critical label imbalance ratio 
      \Function{MultiLabel$\_$SMOTE}{k, D, LIR$_{cr}$}
      \State L $\xleftarrow{}$ set(D.labels)
      \State LIR$_{cr}$ = 2   
      \State upsampled$\_$samples = $[\,\,]$
        \For{$l$ \textbf{in} L}
            \State IR$_{l}$ $\xleftarrow{}$ calculate imbalance ratio (IR) of label $l$
            \While{IR$_{l}$ > LIR$_{cr}$}
                \State set$\_$samples $\xleftarrow{}$ get all samples with label $l$
                \For{$s$ \textbf{in} set$\_$samples}
                    \State set$\_$neighs: $k$ nearest neighbors of $s$ in set$\_$samples
                    \State ref$\_$neigh: a random neigh from set$\_$neighs 
                    \State synth$\_$sample: HELPER($s$,\,ref$\_$neigh,\,set$\_$neighs)
                \EndFor
                \State D.append(synth$\_$sample)
                \State IR$_{l}$ $\xleftarrow{}$ calculate imbalance ratio (IR) of label $l$
            \EndWhile
        \EndFor
        \State \Return D
       \EndFunction
        \Function{helper}{$s$, ref$\_$neigh, set$\_$neighs}
            \State $\triangleright$ finding the features of the synthetic sample
            \For{$f$ \textbf{in} $s$.features}
                \State $\alpha \in [0,1]$ $\xleftarrow{}$ select a random number
                \State synth$\_$sample.$f$ = $f$ + $\alpha$ (ref$\_$neigh.$f$ $-$ $f$)
            \EndFor
            \State $\triangleright$ finding the label of the synthetic sample
            \State synth$\_$sample.label: most frequent label in $s$ and set$\_$neighs
            \State \Return synth$\_$sample 
       \EndFunction
\end{algorithmic}
\label{alg:SMOTE_binary2}
\end{algorithm}

\footnotesize 
\bibliographystyle{unsrt}
\bibliography{references}

\begin{thebibliography}{10}

\bibitem{1347773}
{Min Qin} and {Kai Hwang}.
\newblock Frequent episode rules for internet anomaly detection.
\newblock In {\em Third IEEE International Symposium on Network Computing and
  Applications, 2004. (NCA 2004). Proceedings.}, pages 161--168, 2004.

\bibitem{helmer1998intelligent}
Guy~G Helmer, Johnny~SK Wong, Vasant Honavar, and Les Miller.
\newblock Intelligent agents for intrusion detection.
\newblock In {\em 1998 IEEE Information Technology Conference, Information
  Environment for the Future (Cat. No. 98EX228)}, pages 121--124. IEEE, 1998.

\bibitem{salvador2004learning}
Stan Salvador, Philip Chan, and John Brodie.
\newblock Learning states and rules for time series anomaly detection.
\newblock In {\em FLAIRS conference}, pages 306--311, 2004.

\bibitem{1463166}
C~W Thompson and R~Jena.
\newblock {Digital licensing [software reuse]}.
\newblock {\em IEEE Internet Computing}, 9(4):85--88, jul 2005.

\bibitem{Callister2003}
Paul~D. Callister.
\newblock {Digital Content Licensing}.
\newblock {\em Encyclopedia of Library and Information Science}, pages
  475--483, 2003.

\bibitem{Wang2003}
Xin Wang.
\newblock {Digital rights management for broadband content distribution}.
\newblock {\em Proceedings - 2003 Symposium on Applications and the Internet,
  SAINT 2003}, 21:4, 2003.

\bibitem{Azad2010}
Mir~Mohammad Azad, Abu~Hasnat {Shohel Ahmed}, and Asadul Alam.
\newblock {\em {Digital Rights Management}}, volume~10.
\newblock aug 2010.

\bibitem{Dushkin2019}
Eyal Dushkin, Shay Gershtein, Tova Milo, and Slava Novgorodov.
\newblock {Query driven data labeling with experts: Why pay twice?}
\newblock {\em Advances in Database Technology - EDBT}, 2019-March:698--701,
  2019.

\bibitem{Ratner2016}
Alexander Ratner, Christopher {De Sa}, Sen Wu, Daniel Selsam, and Christopher
  R{\'{e}}.
\newblock {Data programming: Creating large training sets, quickly}.
\newblock {\em Advances in Neural Information Processing Systems},
  (Nips):3574--3582, 2016.

\bibitem{Zhou2018}
Zhi~Hua Zhou.
\newblock {A brief introduction to weakly supervised learning}.
\newblock {\em National Science Review}, 5(1):44--53, 2018.

\bibitem{Sun2017}
Boyuan Sun, Qiang Ma, Shanfeng Zhang, Kebin Liu, and Yunhao Liu.
\newblock {ISelf: Towards cold-start emotion labeling using transfer learning
  with smartphones}.
\newblock {\em ACM Transactions on Sensor Networks}, 13(4), 2017.

\bibitem{Rousseau2011}
Fran{\c{c}}cois Rousseau, Piotr~A. Habas, and Colin Studholme.
\newblock {A Supervised Patch-Based Approach for Human Brain Labeling}.
\newblock {\em IEEE Transactions on Medical Imaging}, 30(10):1852--1862, oct
  2011.

\bibitem{1806.05233}
Soheil Esmaeilzadeh, Yao Yang, and Ehsan Adeli.
\newblock End-to-end parkinson disease diagnosis using brain mr-images by
  3d-cnn, 2018, arXiv:1806.05233.

\bibitem{esm}
Soheil Esmaeilzadeh, Dimitrios~Ioannis Belivanis, Kilian~M. Pohl, and Ehsan
  Adeli.
\newblock End-to-end alzheimer's disease diagnosis and biomarker
  identification.
\newblock In {\em Machine Learning in Medical Imaging}, pages 337--345.
  Springer International Publishing, 2018.

\bibitem{Hakkani-Tur2002}
Dilek Hakkani-T{\"{u}}r, Giuseppe Riccardi, and Allen Gorin.
\newblock {Active learning for automatic speech recognition}.
\newblock {\em ICASSP, IEEE International Conference on Acoustics, Speech and
  Signal Processing - Proceedings}, 4(September 2002), 2002.

\bibitem{Rosenberg2002}
C.~Rosenberg and M.~Hebert.
\newblock {Training Object Detection Models with Weakly Labeled Data}.
\newblock In {\em Procedings of the British Machine Vision Conference 2002},
  pages 56.1--56.10. British Machine Vision Association, 2002.

\bibitem{1904.00788}
Soheil Esmaeilzadeh, Gao~Xian Peh, and Angela Xu.
\newblock Neural abstractive text summarization and fake news detection, 2019,
  arXiv:1904.00788.

\bibitem{Huang2015}
Xin Huang, Chunlei Weng, Qikai Lu, Tiantian Feng, and Liangpei Zhang.
\newblock {Automatic labelling and selection of training samples for
  high-resolution remote sensing image classification over urban areas}.
\newblock {\em Remote Sensing}, 7(12):16024--16044, 2015.

\bibitem{AbdullahAlMamun2018}
S.~M. {Abdullah Al Mamun} and Juha Valimaki.
\newblock {Anomaly detection and classification in cellular networks using
  automatic labeling technique for applying supervised learning}.
\newblock {\em Procedia Computer Science}, 140:186--195, 2018.

\bibitem{Bondi2017}
Elizabeth Bondi, Fei Fang, Debarun Kar, Venil Noronha, Donnabell Dmello, Milind
  Tambe, Arvind Iyer, and Robert Hannaford.
\newblock {VIOLA: Video Labeling Application for Security Domains}.
\newblock {\em Lecture Notes in Computer Science (including subseries Lecture
  Notes in Artificial Intelligence and Lecture Notes in Bioinformatics)}, 10575
  LNCS:377--396, 2017.

\bibitem{Chawla2002}
N.~V. Chawla, K.~W. Bowyer, L.~O. Hall, and W.~P. Kegelmeyer.
\newblock {SMOTE: Synthetic Minority Over-sampling Technique}.
\newblock {\em Journal of Artificial Intelligence Research}, 16(2):321--357,
  jun 2002.

\bibitem{6235959}
G~Ditzler and R~Polikar.
\newblock {Incremental Learning of Concept Drift from Streaming Imbalanced
  Data}.
\newblock {\em IEEE Transactions on Knowledge and Data Engineering},
  25(10):2283--2301, oct 2013.

\bibitem{Charte2015}
Francisco Charte, Antonio~J. Rivera, Mar{\'{i}}a~J. {Del Jesus}, and Francisco
  Herrera.
\newblock {MLSMOTE: Approaching imbalanced multilabel learning through
  synthetic instance generation}.
\newblock {\em Knowledge-Based Systems}, 89:385--397, 2015.

\bibitem{Giraldo-Forero2013}
Andr{\'{e}}s~Felipe Giraldo-Forero, Jorge~Alberto Jaramillo-Garz{\'{o}}n,
  Jos{\'{e}}~Francisco Ruiz-Mu{\~{n}}oz, and C{\'{e}}sar~Germ{\'{a}}n
  Castellanos-Dom{\'{i}}nguez.
\newblock {Managing imbalanced data sets in multi-label problems: A case study
  with the SMOTE algorithm}.
\newblock {\em Lecture Notes in Computer Science (including subseries Lecture
  Notes in Artificial Intelligence and Lecture Notes in Bioinformatics)}, 8258
  LNCS(PART 1):334--342, 2013.

\bibitem{1437839}
Y~Wang, J~Wong, and A~Miner.
\newblock {Anomaly intrusion detection using one class SVM}.
\newblock In {\em Proceedings from the Fifth Annual IEEE SMC Information
  Assurance Workshop, 2004.}, pages 358--364, jun 2004.

\bibitem{Karczmarek2020}
Pawe{\l} Karczmarek, Adam Kiersztyn, Witold Pedrycz, and Ebru Al.
\newblock {K-Means-based isolation forest}.
\newblock {\em Knowledge-Based Systems}, 195:105659, 2020.

\bibitem{Antonini2018}
Mattia Antonini, Massimo Vecchio, Fabio Antonelli, Pietro Ducange, and Charith
  Perera.
\newblock {Smart audio sensors in the internet of things edge for anomaly
  detection}.
\newblock {\em IEEE Access}, 6:67594--67610, 2018.

\bibitem{Ma2016}
Mathew~X. Ma, Henry~Y.T. Ngan, and Wei Liu.
\newblock {Density-based outlier detection by local outlier factor on
  largescale traffic data}.
\newblock {\em IS and T International Symposium on Electronic Imaging Science
  and Technology}, pages 1--4, 2016.

\bibitem{Cheng2019}
Zhangyu Cheng, Chengming Zou, and Jianwei Dong.
\newblock {Outlier detection using isolation forest and local outlier}.
\newblock {\em Proceedings of the 2019 Research in Adaptive and Convergent
  Systems, RACS 2019}, pages 161--168, 2019.

\bibitem{Bi2005}
Jinbo Bi and Tong Zhang.
\newblock {Support vector classification with input data uncertainty}.
\newblock {\em Advances in Neural Information Processing Systems}, 2005.

\bibitem{7898482}
S~Zhang, X~Li, M~Zong, X~Zhu, and R~Wang.
\newblock {Efficient kNN Classification With Different Numbers of Nearest
  Neighbors}.
\newblock {\em IEEE Transactions on Neural Networks and Learning Systems},
  29(5):1774--1785, may 2018.

\bibitem{Sharma2016}
Himani Sharma and Sunil Kumar.
\newblock {A Survey on Decision Tree Algorithms of Classification in Data
  Mining}.
\newblock {\em International Journal of Science and Research (IJSR)},
  5(4):2094--2097, 2016.

\bibitem{Cutler2012}
L~Breiman.
\newblock {Random forests}.
\newblock {\em Machine learning}, pages 157--175, 2001.

\bibitem{Mason2000}
Llew Mason, Jonathan Baxter, Peter Bartlett, and Marcus Frean.
\newblock {Boosting algorithms as gradient descent}.
\newblock {\em Proceedings of the 12th International Conference on Neural
  Information Processing Systems}, pages 512--518, 2000.

\bibitem{939503}
Y~L Murphey, {Zhihang Chen}, and {Hong Guo}.
\newblock {Neural learning using AdaBoost}.
\newblock In {\em IJCNN'01. International Joint Conference on Neural Networks.
  Proceedings (Cat. No.01CH37222)}, volume~2, pages 1037--1042 vol.2, jul 2001.

\bibitem{Tibshirani2003}
Robert Tibshirani, Trevor Hastie, Balasubramanian Narasimhan, and Gilbert Chu.
\newblock {Class Prediction by Nearest Shrunken Centroids, with Applications to
  DNA Microarrays}.
\newblock {\em Statistical Science}, 18(1):104--117, feb 2003.

\bibitem{4038449}
T~M Cover.
\newblock {Geometrical and Statistical Properties of Systems of Linear
  Inequalities with Applications in Pattern Recognition}.
\newblock {\em IEEE Transactions on Electronic Computers}, EC-14(3):326--334,
  jun 1965.

\bibitem{Abramson1963}
N.~Abramson, D.~Braverman, and G.~Sebestyen.
\newblock {\em {Pattern recognition and machine learning}}, volume~9.
\newblock 1963.

\bibitem{Ebden2008}
M~Ebden.
\newblock {Gaussian Processes for Regression: A Quick Introduction}.
\newblock 2008.

\bibitem{Zhi2002}
Xiaojin Zhi and Zoubin Ghahremani.
\newblock {Learning from Labeled and Unlabeled Data with Label Propagation},
  2002.

\bibitem{xgboost}
Tianqi Chen and Carlos Guestrin.
\newblock Xgboost: A scalable tree boosting system.
\newblock In {\em Proceedings of the 22nd ACM SIGKDD International Conference
  on Knowledge Discovery and Data Mining}, KDD '16, page 785–794, New York,
  NY, USA, 2016. Association for Computing Machinery.

\end{thebibliography}


\end{document}